\documentclass[10pt,conference,a4paper]{IEEEtran}
\usepackage{amsmath}
\usepackage{array}
\usepackage{url}
\usepackage{graphicx}
\usepackage{amsmath,amssymb} % define this before the line numbering.
\usepackage{color}
\usepackage[utf8]{inputenc}
\usepackage{version}
\usepackage[tight, footnotesize]{subfigure}
\usepackage{relsize}
\usepackage{paralist}
\usepackage[super]{nth}
\usepackage{capt-of,etoolbox}
\usepackage{multirow}
\usepackage{xcolor}
\usepackage[ruled, vlined]{algorithm2e}
\usepackage[textwidth=3.6cm, textsize=footnotesize]{todonotes}
\DeclareMathOperator{\argmin}{argmin}

\newcommand{\GG}{\mathcal{G}}
\newcommand{\SSS}{\mathcal{S}}
\newcommand{\CC}{\mathcal{C}}
\newcommand{\PP}{\mathcal{\mathbf{P}}}
\newcommand{\eqsize}{\small}

\begin{document}
\title{SCALP: Superpixels with Contour Adherence \\using Linear Path}

\author{\IEEEauthorblockN{Rémi Giraud\textsuperscript{1,2,3}, Vinh-Thong Ta\textsuperscript{1,2} and Nicolas Papadakis\textsuperscript{3}}
\IEEEauthorblockA{\small 
\textsuperscript{1}Univ. Bordeaux, CNRS, LaBRI, UMR 5800, PICTURA, F-33400, Talence, France\\
\textsuperscript{2}Bordeaux INP, LaBRI, UMR 5800, F-33400, Talence, France\\
\textsuperscript{3}Univ. Bordeaux, CNRS, IMB, UMR 5251, F-33400, Talence, France\\
Email: (remi.giraud, vinh-thong.ta)@labri.fr, nicolas.papadakis@math.u-bordeaux.fr} 
}

\maketitle

\begin{abstract}
Superpixel decomposition methods are generally used as a pre-processing step to speed up image processing tasks.
They group the pixels of an image into homogeneous regions while trying to respect existing contours. 
For all state-of-the-art superpixel decomposition methods, a trade-off is made between 1)
computational time,
2) adherence to image contours and 3)
regularity and compactness of the decomposition.
In this paper, we propose a fast method to compute Superpixels with Contour Adherence using Linear Path (SCALP)
in an iterative clustering framework. 
The distance computed when trying to associate a pixel to a superpixel during the clustering is
enhanced by considering
the linear path to the superpixel barycenter.
The proposed framework produces regular and compact superpixels that adhere to the image contours.
We provide a detailed evaluation of SCALP on the standard Berkeley Segmentation Dataset.
The obtained results outperform state-of-the-art methods
in terms of standard superpixel and contour detection metrics.

\end{abstract}

\IEEEpeerreviewmaketitle

\section{Introduction}

Image segmentation is an essential tool to analyze the image content.
The aim is to split the
image into similar regions
with respect to some priors ({\emph{e.g.}, object, color or texture).
To decrease the computational time and to
improve the accuracy of unsupervised segmentation, 
superpixel decomposition methods
have been proposed.
These methods group the pixels into homogeneous regions
while trying to respect image contours. 
Superpixels drastically decrease the image content dimension
while preserving geometrical information, 
contrary to multi-resolution approaches.
For instance, small objects that disappear at small resolution levels, 
can still be contained into single superpixels.
Hence, superpixels have naturally become 
key building blocks of many computer vision works such as:
 contour detection \cite{arbelaez2011}, face labeling \cite{kae2013},
 object localization \cite{fulkerson2009},
 or
multi-class object segmentation
\cite{gould2008,gould2014}.

Superpixel methods can be divided in two categories that provide either irregular or regular decompositions.
With irregular methods,
superpixels can be stretched, with very different sizes, 
and 
may
overlap with several objects contained in the image.
Moreover, very small superpixels can be extracted, and without enough pixels,
significant descriptors are difficult to estimate.
On the contrary, regular methods provide 
superpixels with approximately the same size,
and enable to compute more robust descriptors.

There is no general rule for the definition of an optimal superpixel method
since the desired properties depend on the tackled application.
A compromise must be made between:
computational time,
adherence to image boundaries, and
size and shape regularity of the superpixel decomposition.

On the one hand, accurate contour adherence is for instance reached with irregular methods (\emph{e.g.}, \cite{vandenbergh2012})
that allow very stretched superpixel decompositions based on color similarity.
On the other hand, it appears crucial for
superpixel-based object recognition methods 
to use regular decompositions \cite{gould2008,gould2014}. 
Moreover, when facing  video tracking of superpixels \cite{wang2011},
fast and regular approaches are suitable to consider the time consistency of the decomposition. 
Regularity is thus crucial to accurately analyze object trajectories in a scene.
In Figure \ref{fig:reg}, we show an example of a reconstructed image 
from the irregular method SEEDS \cite{vandenbergh2012}
and from our regular method SCALP.
The image is computed by the average color on each superpixel.
SCALP provides a much more visually satisfying result due to regularity and accuracy of superpixel  boundaries.

\newcommand{\wo}{0.155\textwidth}
\newcommand{\ho}{0.23\textwidth}
\begin{figure}[h!]
\centering
{\scriptsize
\begin{tabular}{@{}c@{\hspace{1mm}}c@{\hspace{1mm}}c@{}}
\includegraphics[height=\ho,width=\wo]{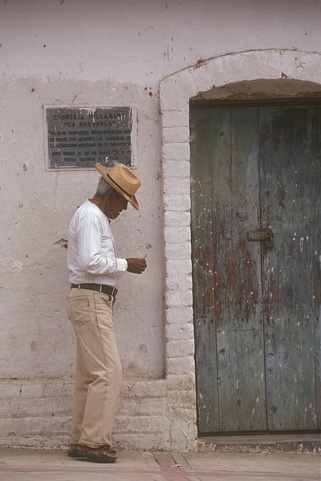}&
\includegraphics[height=\ho,width=\wo]{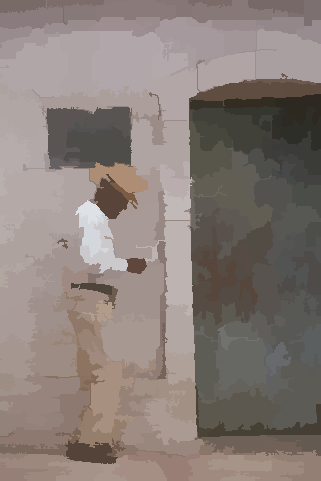}&
\includegraphics[height=\ho,width=\wo]{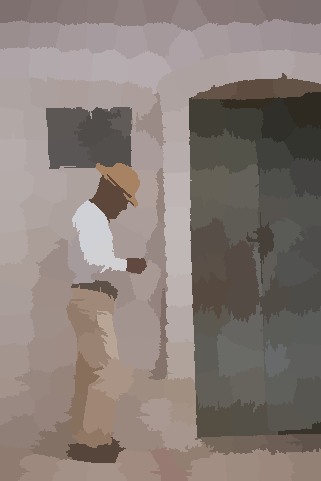}\\
Initial image & SEEDS \cite{vandenbergh2012} & SCALP 
\end{tabular}
}
\caption{Reconstruction from average colors with 200 superpixels on an example image 
from the Berkeley Segmentation Dataset \cite{martin2001}
}
\label{fig:reg}
\end{figure}

\textbf{Irregular Superpixel Methods.} 
Classical methods, such as the watershed approach \cite{vincent91},
compute decompositions of highly irregular size and shape.
In this context, starting from an initial clustering, Mean shift \cite{comaniciu2002} or Quick shift \cite{vedaldi2008} approaches 
use histogram-based
segmentation but require high computational time. 
By considering pixels as nodes of a graph,  faster agglomerative clustering can be obtained \cite{felzenszwalb2004}.
In addition to the lack of control over the superpixels shape, all these methods present another main drawback. 
They do not allow to directly control the number of superpixels,
which is a major issue when using superpixels as low-level representation
to reduce the computational time for a dedicated task.

More recently, the SEEDS method \cite{vandenbergh2012} has been proposed to produce  
a decomposition in a reduced computational time.
This approach is initialized with a regular grid and updates superpixel boundaries
with block and pixel transfers.
Despite its initial regular grid, 
this method provides superpixels with irregular shapes.
Finally, the authors report significantly degraded results when trying to regularize the superpixel shape with  
compactness constraint \cite{vandenbergh2012}.

\textbf{Regular Superpixel Methods.} 
When considering more general applications than contour adherence, 
state-of-the-art superpixel-based methods consider regular decompositions (\emph{e.g.}, \cite{wang2011,gould2014}).
Classical methods are based on region growing, such as Turbopixels \cite{levinshtein2009} using geometric flows,
or graph-based energy models \cite{veksler2010,liu2011}.
In \cite{machairas2015}, the watershed method is adapted to produce regular decompositions using
a spatially regularized gradient.

Recently, the Simple Linear Iterative Clustering (SLIC) superpixel method was proposed in \cite{achanta2012}, 
and its extensions, \emph{e.g.}, \cite{li2015,zhang2016}.
This method performs an accurate color clustering, providing regular superpixels,
while being order of magnitude faster than graph approaches \cite{veksler2010} or \cite{liu2011}, and achieves state-of-the-art results
on superpixel metrics. 
However, since a compactness parameter is set to enforce the superpixel shape regularity, 
SLIC can fail to adhere to image contours, as for other regular methods \cite{levinshtein2009,veksler2010}.

Several works have attempted to enhance the performance of regular methods in terms of contour adherence by using a contour prior information.
Although methods to compute regions from contours have been proposed (\emph{e.g.}, \cite{arbelaez2009}), 
they do not allow to control the size, shape and number of regions, and therefore cannot be considered 
as superpixel decomposition methods.
In \cite{mori2004}, contour priors are used to compute a pre-segmentation 
using the normalized cuts algorithm \cite{shi2000}. This segmentation is considered as a hard constraint 
to provide a finer regular decomposition.
However, the decompositions based on normalized cuts are highly tuned and computationally expensive, while they are far from state-of-the-art results in terms of contour adherence.
In \cite{moore2008}, the decomposition is constrained to fit to a grid, 
 named superpixel lattice. This decomposition uses a contour map as input  
to determine this lattice, and iteratively refines it using optimal cuts.
The method finally produces superpixels of regular sizes but irregular shapes, that are visually unsatisfactory \cite{moore2008}.
Moreover, the method appears very dependent on the contour prior. 
In more recent works, such as \cite{zhang2016}, superpixels are locally forced to adhere to 
contours by considering a pre-computed gradient 
used as constraint to compute the boundaries of superpixels.

\textbf{Contributions.} 
In this paper, we propose a fast method to directly
include a contour prior in a superpixel clustering framework,
and not as hard prior from a pre-segmentation step.
To that end, the distance computed when trying to associate a pixel to a superpixel is
enhanced by considering image feature and contour intensity on a linear path to the superpixel barycenter.
The decomposition provides superpixels of regular sizes and shapes 
that respect color homogeneity, and their
boundaries are computed according to the contour prior.

%%%%%%%%%%%%%%%%%%%%%%
We provide a detailed evaluation of our method on the standard Berkeley Segmentation Dataset \cite{martin2001},
compared to state-of-the-art methods on superpixel and contour detection metrics.
We demonstrate the regularity of our decomposition, that obtains
the best results on most of the compared metrics.

\section{SCALP Framework}

The SCALP framework generalizes the iterative clustering algorithm of \cite{achanta2012}. 
Thanks to the introduction of the linear path within the clustering,
a more regular decomposition can be obtained. 
Moreover, prior information such as contour maps can be naturally included within 
the path  to softly constrain the decomposition, as illustrated in Figure \ref{fig:scap}.

In this section, we first present the iterative clustering framework. 
Then, we define the linear path to the superpixel barycenter.
Next, we propose to use this path to include relevant information for superpixel clustering, 
by proposing a new color distance term.
We finally propose a fast method to integrate a contour prior into our framework.

\begin{figure}[h!]
\centering
\includegraphics[height=0.21\textwidth,width=0.48\textwidth]{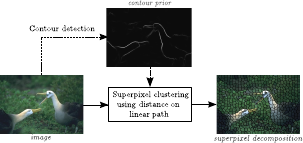}
\caption{The SCALP framework. 
When trying to associate a pixel to a superpixel,
SCALP considers the linear path to the superpixel barycenter 
with a color distance.
A prior can be used (dotted arrows) to ensure 
that no image contour is crossed, leading to an accurate and homogeneous decomposition
}
\label{fig:scap}
\end{figure}

\subsection{Simple Linear Iterative Clustering}

As previously stated, 
 SLIC \cite{achanta2012} 
is one of the most efficient and simplest superpixel frameworks.
In its default settings, it only takes the number of superpixels  as parameter.
The decomposition is initialized with
a regular grid, with blocks of size $r$,
and an iterative clustering, spatially constrained into a window of fixed size $2r+1$,
is performed for all superpixels.
The size $r$ is defined by the ratio between $N$ the number of pixels and $k$
the number of superpixels, such that $r=\sqrt{N/k}$.
With this constraint, roughly equally sized superpixels are provided, 
ensuring the decomposition regularity.

Each superpixel $S_k$ of SLIC is described by a cluster $C_k$ containing 
the average feature information and spatial barycenter of all pixels $p\in S_k$.
In \cite{achanta2012}, the clustering is performed in the CIELab color space. 
Therefore, each superpixel $S_k$ at a given iteration is described by its cluster $C_k$=$[l_k,a_k,b_k,X_k]$, 
containing the average CIELab color feature on pixels $p\in S_k$ and
$X_k$=$[x_k,y_k]$, the barycenter of $S_k$.
At each iteration, and for each cluster $C_k$,
all pixels $p$=$[l_p,a_p,b_p,X_p]$,  within a square window of size $2r+1{\times}2r+1$
centered on the barycenter $X_k$,
are tested to be associated to $S_k$ by computing a spatial distance 
$d_s(p,C_k) = {(x_p-x_k)^2 + (y_p - y_k)^2},$ and a color distance
$d_c(p,C_k) = {(l_p-l_k)^2 + (a_p-a_k)^2 + (b_p-b_k)^2}.$
The pixel $p$ is assigned to the cluster $C_k$ that minimizes the sum of these two distances.
Despite its basic framework,
SLIC achieves results that are comparable to state-of-the-art methods and even 
outperforms irregular approaches such as \cite{vandenbergh2012}, in the contour detection evaluation framework of \cite{martin2004}.

The framework in \cite{achanta2012} provides regular superpixels that can adhere to image contours.
However, this adherence can still be enhanced 
by considering accurate contour information as prior.
By introducing the notion of linear path, SCALP computes a generalized color distance term that 
improves 
the homogeneity of the color clustering and
the regularity of the superpixels shape. 
We also propose to integrate
the contour prior information as a soft constraint in this new color distance
to enforce the adherence to image contours.

\subsection{Linear Path to the Cluster Center}

To enforce the color homogeneity within a superpixel and contour adherence,
we
consider the color and contour intensities  
on the linear path between pixels and their corresponding superpixel barycenter.

We thus define the path $\PP^k_p$ of pixels between a pixel $p$ and a superpixel barycenter $X_k$.
By considering information along this path, SCALP is able to enhance the relevance of the color distance.
Note that the more pixels are considered in $\PP^k_p$, the higher the computational cost is.
Hence, we propose to use
\cite{bresenham1965} 
to only get the positions of pixels on the direct path between 
the pixel $p$ and the barycenter of superpixel $S_k$, as illustrated in Figure \ref{fig:bres_diag}.
The considered pixels $q$ (in red) are those that intersect with the segment (red arrow) between $X_p$,
the position of pixel $p$ (in black),
and $X_k$, the barycenter of superpixel $S_k$ (in blue).
By considering this simple linear path instead of a more sophisticated  geodesic one \cite{zeng2011}, we limit the computational cost and enforce 
the decomposition compactness.

\begin{figure}[h!]
\centering
\includegraphics[height=87pt,width=0.42\textwidth]{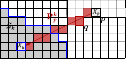}
\caption{Illustration of linear path to cluster center
(see text for more details)
}
\label{fig:bres_diag}
\end{figure}

\subsection{Improved Color Distance to Cluster}

As in \cite{achanta2012}, the spatial distance $d_s$
between a processed pixel $p$ and a cluster $C_k$
is only computed between positions of $p$ and its barycenter $X_k$.
However, the CIELab color distance $d_c$
may now be computed on $\PP^k_p$, the set of pixels 
on the path to the superpixel barycenter. 
The new color distance is thus defined as:
\begin{equation}
\label{newdist0}d_c(p,C_k,\PP^k_p )=\lambda d_c(p,C_k) + (1-\lambda)\frac{1}{|\PP^k_p|}\hspace{-0.1cm}\sum_{q\in \PP^k_p}\hspace{-0.05cm}d_c(q,C_k),
\end{equation}
where $\lambda\in[0,1]$ weights the influence of the color distance along the path between $X_p$ and $X_k$. 
Since color on the linear path to the barycenter should be close to the average color of the superpixel,
SCALP naturally enforces the decomposition compactness and favors uniform color distribution.

\subsection{Adherence to Contour Prior}
When associating a pixel to a superpixel cluster, we want to favor the color homogeneity, 
the proximity to the cluster barycenter and the adherence to image contours. 
We assume that a soft contour prior map $\CC$ is available. 
Such map typically sets $\CC(p)\rightarrow1$
if a contour is detected,
otherwise $\CC(p)\rightarrow0$,
at pixel $p$. 
A fast and efficient way to integrate this prior information is to weight the color distance \eqref{newdist0} by  $d_{\CC}(p,C_k,\PP^k_p)$, 
the sum of contour intensity on $\PP^k_p$ defined as:
\begin{equation}
\label{newweight}d_{\CC}(p,C_k,\PP^k_p) = 1+\frac{\gamma}{|\PP^k_p|}\sum_{q\in \PP^k _p}{\left(1 - \exp(-\CC(q)^2/\sigma^2)\right)},  
\end{equation}
where $\gamma$ $\geq$ $0$ and $\sigma$ $>$ $0$ are parameters that weight the influence
of the contour prior along the linear path.
When a contour intersects the path between a pixel $p$ and a cluster barycenter, 
such term tends to prevent this pixel 
to be associated to the cluster $C_k$.
The proposed distance $D$ to minimize during the clustering is finally defined as:
\begin{align}
\hspace{-0.2cm} D(p,C_k) &= d_c(p,C_k,\PP^k_p)d_{\CC}(p,C_k,\PP^k_p) + d_s(p,C_k)\frac{m^2}{r^2} , \label{newdist}
\end{align}
where $m$ is the compactness parameter, \emph{i.e.}, setting 
the trade-off between color distance $d_c$ and spatial distance $d_s$.
The higher $m$ is, the more regular, \emph{i.e.}, compact, is the superpixel shape. On the other hand, small values of $m$
allow a better adherence to image color boundaries,
producing superpixels of more variable sizes and shapes.
By setting $\lambda=1$ in \eqref{newdist0} and $\gamma=0$ in \eqref{newweight},
the proposed distance \eqref{newdist} reduces to 
the standard distance used in \cite{achanta2012}.
The SCALP algorithm is summarized in Algorithm \ref{SCALP}.
\textit{Remark:}
Note that although the algorithm starts from an initial regular grid,
the spatial barycenter of
a superpixel 
may, in very rare cases,
fall outside the superpixel after a few iterations.
Having a cluster center outside the superpixel impacts 
the computation of the linear path.
Hence, if the barycenter $X_k$ is not contained into the superpixel $S_k$,  
we consider the projected position $X_k^*$ to compute the linear path to the center $\PP^k_p$,
which is computed as:
\begin{equation}
{X_k^*} = \underset{{X_p}, p\in S_k}{\argmin} \|{X_p} - {X_k}\|_2. \label{center}
\end{equation}

\begin{algorithm}
\footnotesize
\SetKwInOut{Input}{inputs}\SetKwInOut{Output}{output}

\Input{Number of superpixels $k$, Contour prior $\mathcal{C}$}
\Output{Superpixel decomposition $\SSS$}
Initialization of clusters $C_k \gets [l_k,a_k,b_k,X_k]$ from a regular grid\\
Initialization of superpixel labels $\SSS \gets 0$

\For{\emph{each iteration}}{
  Distance $d \gets \infty$\\
  \For{\emph{each $C_k$}}{
    \For{\emph{each $p$ in a $2r+1{\times}2r+1$ window centered on $X_k$}}{
      Compute $D(p,C_k)$ using $\mathcal{C}$ and $\PP_p^k$
      \eqref{newdist}\\
      \If{$D(p,C_k) < d(p)$}{
        $d(p) \gets D(p,C_k)$  and  $\SSS(p) \gets k$
      }
    }
  }
  Update $[l_k,a_k,b_k,X_k]$ for all clusters $C_k$\\
  Compute if necessary the projection $X_k^*$ of centers $X_k$ into $S_k$ \eqref{center}
}
\caption{SCALP method}\label{SCALP}
\end{algorithm}

\subsection{Contour Prior}

SCALP can directly consider a contour prior into the clustering framework.
Therefore, our decomposition is not constrained by a pre-segmentation step.
This contour prior can either be soft or hard, \emph{i.e.}, having values between 0 and 1 or 
being binary, and can be computed by any contour detection method (see for instance \cite{arbelaez2011} and references therein).
In Section \ref{influ_params}, we report results obtained using different contour detection methods \cite{maire2008}, \cite{xiaofeng2012}, \cite{dollar2013}.

\section{Results}

\subsection{\label{valid}Validation Framework}

\subsubsection{Dataset}

We evaluate our method on the standard Berkeley Segmentation Dataset (BSD) \cite{martin2001}.
The BSD contains 500 images of $321{\times}481$ pixels divided into three sets: 
200 for training, 100 for validation and 200 for testing.
For each image, at least 5 ground truth decompositions from manual segmentations are also provided
to compute evaluation metrics.
We report results of SCALP and other compared methods on the 200 images of the test set.

\subsubsection{Metrics}

To evaluate the performances of our framework and compare to the state-of-the-art methods, 
we use the standard superpixel evaluation metrics: boundary recall (BR), undersegmentation error (UE) and
achievable segmentation accuracy (ASA).
To compare the regularity of superpixel shape of different decompositions,
we report the compactness measure introduced in \cite{schick2012}.
In the following, for an image $I$, a human ground truth segmentation
is denoted $\GG=\{G_i\}_{i\in \{1,\dots ,n_G\}}$, where $G_i$ is a  segmented region of the scene,   
$n_G$ is the number of regions within the decomposition, and $|.|$
denotes the cardinality.
Reported values are averaged results on all
ground truths.
To quantify the results of contour detection performance,
we
report the precision (P) recall (R) curves \cite{martin2004}.

\noindent\textbf{Boundary Recall.} 
This measure evaluates the percentage of ground truth contours $\mathcal{B(\GG)}$ that overlap, 
within an $\epsilon$-pixel distance,
with the boundaries of the computed superpixel decomposition $\mathcal{B}(\SSS)$. The BR metric is defined as follows:

{\eqsize
\begin{equation}
BR(\SSS,\GG) = \frac{\sum_{p\in\mathcal{B}(\GG)}\delta[\min_{q\in\mathcal{B}(\SSS)}\|p-q\|< \epsilon]}{|\mathcal{B}(\GG)|}  ,   \label{br}
\end{equation}
}\noindent
with $\delta[a]$=$1$ when $a$ is true and $0$ otherwise,
and $\epsilon$=$2$ as in \cite{vandenbergh2012}.

\noindent\textbf{Achievable Segmentation Accuracy.} 
The ASA is an upper bound measure that computes
the maximum object segmentation accuracy by taking superpixels as units.
For each superpixel $S_k$, the largest possible overlap with a 
ground truth segment $G_i$ is computed and averaged as follows:

{\eqsize
\begin{equation}
ASA(\SSS,\GG) = \frac{\sum_{k}\max_i|S_k\cap G_i|}{\sum_{i}|G_i|}.  \label{asa}
\end{equation}
}

\noindent\textbf{Undersegmentation Error.} 
This measure evaluates the percentage of pixels that cross
ground truth boundaries. 
With an accurate superpixel decomposition with respect to a given ground truth, 
superpixels should overlap with only one object.
The undersegmentation error (UE) is computed as:

{\eqsize
\begin{equation}
UE(\SSS,\GG) = \frac{\sum_{i}\sum_{k:S_k\cap G_i\neq\emptyset}{|S_k-G_i|}}{\sum_{i}|G_i|}.  \label{ue}
\end{equation}
}

\noindent\textbf{Compactness measure.} 
The compactness (CO) measure for a superpixel decomposition $\SSS$
is defined as in \cite{schick2012}: %\vspace{-0.2cm}

{\eqsize
\begin{equation}
CO(\SSS) = \frac{1}{|I|}\sum_{S_k\in\SSS}\frac{4\pi |S_k|^2}{|P(S_k)|^2} ,
 \label{regu_metrics}
\end{equation}
}\noindent
where $P(S_k)$ defines the perimeter of the superpixel $S_k$.
High values indicates more compact
superpixels.

\noindent\textbf{Precision-Recall.} 
The PR framework
\cite{martin2004} 
evaluates the contour detection accuracy.
It can be used to measure the contour detection performances of segmentation or superpixel algorithms.
The PR curves are computed from a set of input maps, which values represent the confidence in being on an image  boundary.
When evaluating superpixel methods, these maps can be
computed by averaging superpixel boundaries
obtained from decompositions at multiple scales. 
As in \cite{vandenbergh2012}, we considered 12 scales, 
ranging from 6 to 600 superpixels,
to compute the boundary maps.
We rank the methods according to their maximum $F$-measure defined as
$(2.P.R)/(P+R)$,
where $P$ (precision) is the percentage of accurate detection among the computed contours,
and $R$ (recall) is the percentage of detected ground truth contours.

\subsubsection{\label{param_settings}Parameter Settings}

SCALP was implemented with MATLAB using C-MEX code,
on a standard Linux computer.
The number of clustering iterations is set to $5$.
We set $\lambda$ to $0.5$ in the proposed color distance \eqref{newdist0}, 
$\sigma$ to $0.25$ in \eqref{newweight}, and $\gamma$ to $2r$, 
to adjust to the superpixel size. 
The compactness parameter $m$ is set to $10$ in \eqref{newdist} as in \cite{achanta2012}.
These parameters offer a good trade-off between adherence to contour prior and compactness.
In the following, if not mentioned, we use a prior from the contour detection method of \cite{dollar2013}.

\setcounter{figure}{4}
\newcommand{\lllh}{0.22\textwidth}
\newcommand{\llll}{0.235\textwidth}
\begin{figure*}[ht!]
\centering
{\footnotesize
\begin{tabular}{c@{\hspace{2mm}}c@{\hspace{2mm}}c@{\hspace{2mm}}c}
\hspace{-0.4cm}
\includegraphics[height=\lllh,width=\llll]{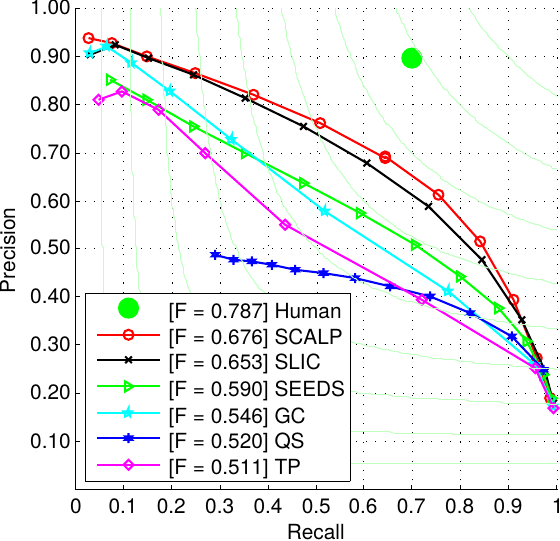}&
\includegraphics[height=\lllh,width=\llll]{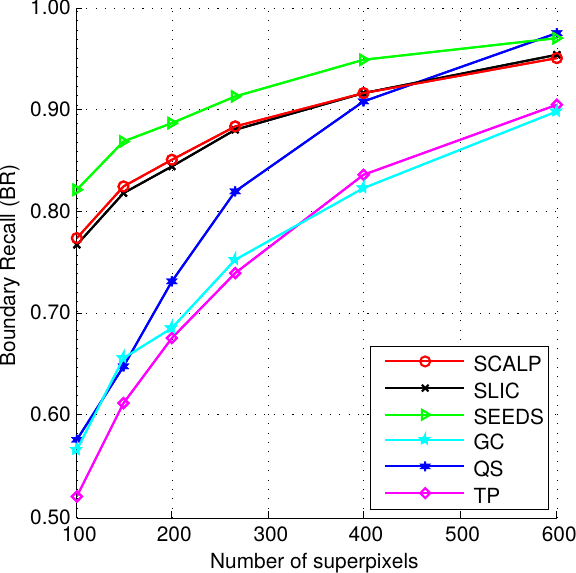}&
\includegraphics[height=\lllh,width=\llll]{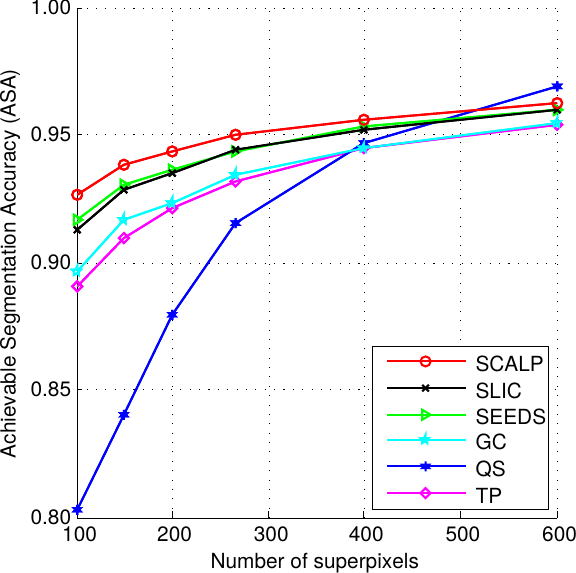}&
\includegraphics[height=\lllh,width=\llll]{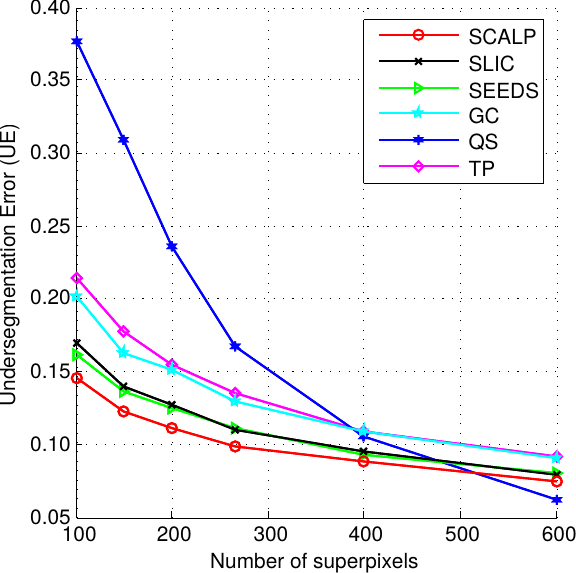}\\
\end{tabular}
}
\caption{Comparison between the proposed SCALP framework and state-of-the-art methods 
on contour detection and superpixel metrics on the BSD test set}
\label{fig:soa_metrics}
\end{figure*}

\subsection{\label{influ_params}Influence of Parameters}

We first measure the influence of parameters within the proposed framework.
In Figure \ref{fig:contour}(a), we provide PR curves for different distance settings.
Contributions of the new the color distance \eqref{newdist0} and the additional contour intensity \eqref{newweight} 
computed on the linear path to the superpixel barycenter both
increase the accuracy of the decomposition with respect to ground truth segmentations.
The complete SCALP algorithm, \emph{i.e.}, using color distance and 
contour intensity, provides the best results and outperforms the standard method proposed in \cite{achanta2012}.

\newcommand{\hw}{0.46\textwidth}
\newcommand{\wh}{0.32\textwidth}
\newcommand{\hhw}{160pt}
\newcommand{\ww}{0.23\textwidth}
\newcommand{\hh}{100pt}
\newcommand{\hhi}{75pt}

\setcounter{figure}{3}
\newcommand{\www}{0.23\textwidth}
\newcommand{\hhh}{0.23\textwidth}
\begin{figure}[t]
\centering
{\footnotesize
\begin{tabular}{c@{\hspace{2mm}}c}
\includegraphics[height=\hhh,width=\www]{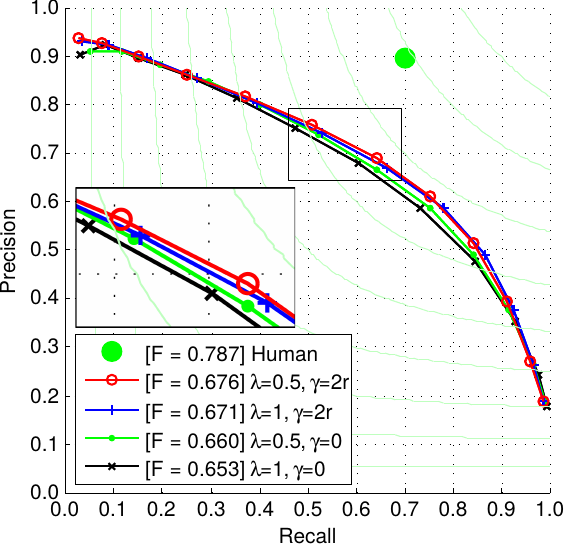}&
\includegraphics[height=\hhh,width=\www]{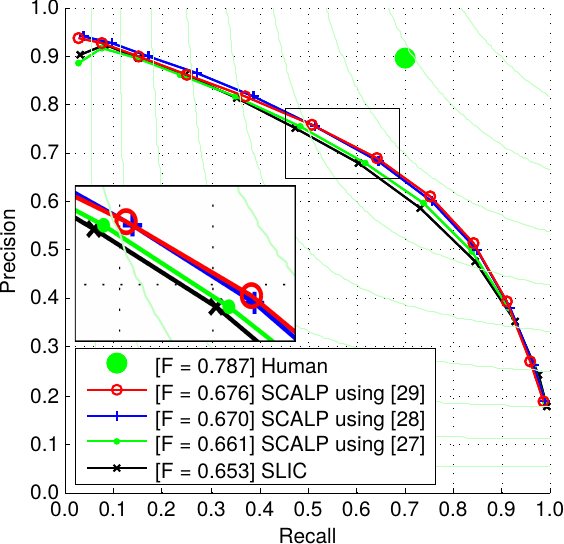}\\
(a)&(b)
\end{tabular}
}
\caption{PR curves for different SCALP distance parameters (a), and
contour detection methods (b)
}
\label{fig:contour}
\end{figure}

We also investigate the influence of the contour prior.
In Figure \ref{fig:contour}(b), we provide PR curves obtained by using
the \textit{globalized probability of boundary}
algorithm \cite{maire2008}, a method using
learned sparse codes of patch gradients \cite{xiaofeng2012}, 
and a structured forests approach
 \cite{dollar2013} for contour detection.
Results are also improved with respect to the contour detection accuracy.
In the following, results are computed using \cite{dollar2013}.

\setcounter{figure}{5}

\subsection{{\label{soa}}Comparison with State-of-the-Art Methods}

We compare the proposed approach to the following state-of-the-art methods:
Quick shift (QS) \cite{vedaldi2008},
Turbopixels (TP) \cite{levinshtein2009},
a graph cut approach (GC) \cite{veksler2010},
SLIC \cite{achanta2012} and
SEEDS \cite{vandenbergh2012}.
Note that TP and GC enforce regularity and thus provide very consistent decompositions
at the expense of lower contour adherence.
All provided results are computed with the same validation framework, described in Section \ref{valid},
with codes provided by the authors and used with their default settings.

In Figure \ref{fig:soa_metrics},
we provide PR curves with their  maximum $F$-measure,
and report the standard BR \eqref{br}, ASA \eqref{asa} and UE
\eqref{ue} metrics. 
SCALP is ranked $\nth{1}$ on PR, providing the higher $F$-measure (0.676).
On superpixel metrics, it is ranked $\nth{1}$ on ASA and UE, and $\nth{2}$ on BR.

Although high BR results indicate that ground truth boundaries are well detected by
the superpixels, this measure does not consider the false detection. 
Therefore, irregular methods such as \cite{vandenbergh2012}, that produces very stretched superpixels with much more boundary pixels 
can obtain higher BR results.
Better PR performances indicate that SCALP very accurately detects object contours with a lower false detection rate.
The UE metric is penalized when superpixels overlap with multiple objects.
Hence, SCALP superpixels overlap with a lower number of ground truth segments.
The ASA evaluates the consistency of a decomposition with respect to
the objects within an image, thus enhancing the largest possible overlap.
Higher ASA results for SCALP also indicate that the produced superpixels are better contained in the image objects.

The regularity of the proposed SCALP framework is confirmed with Table \ref{table:std}, which reports the compactness measure
\eqref{regu_metrics} for the best compared methods.
The provided results are averages obtained on all image decompositions, on the same scales
as the ones used to compute the PR curves.
SCALP obtains the most regular superpixels, even improving the results of  SLIC \cite{achanta2012}.

\begin{table}[ht!]
\renewcommand{\arraystretch}{1.3}
\centering
\caption{Comparison of superpixel regularity, CO
  \eqref{regu_metrics}}
{\small
 \begin{tabular}{c@{\hspace{3mm}}@{\hspace{3mm}}c@{\hspace{3mm}}@{\hspace{3mm}}c@{\hspace{3mm}}@{\hspace{3mm}}c@{\hspace{3mm}}}
 \cline{1-4}
 SEEDS \cite{vandenbergh2012}& QS \cite{vedaldi2008} & SLIC \cite{achanta2012} & SCALP  \\
 \hline 
$0.201$&$0.205$&$0.269$&$\mathbf{0.278}$\\
  \end{tabular}
 }
 \label{table:std}
\end{table}

Finally, Figure \ref{fig:soa_images} illustrates the decomposition 
results obtained with SCALP and the compared approaches on example images. 
SCALP appears to provide more regular superpixels while tightly following the image contours.

SCALP achieves the best state-of-the-art segmentation and contour detection performance, 
while providing a regular superpixel decomposition,
in a limited computational time, \emph{i.e.}, less than $0.4$s per image of the BSD.

\newcommand{\ppp}{64.5pt}
\newcommand{\wwwh}{96.5pt}
\begin{figure*}[t!]
\centering
{\footnotesize
\begin{tabular}{c@{\hspace{2mm}}c@{\hspace{2mm}}c@{\hspace{2mm}}c@{\hspace{2mm}}c}
QS \cite{vedaldi2008}& GC \cite{veksler2010}& SEEDS \cite{vandenbergh2012}&SLIC \cite{achanta2012}& SCALP\\
\includegraphics[height=\ppp,width=\wwwh]{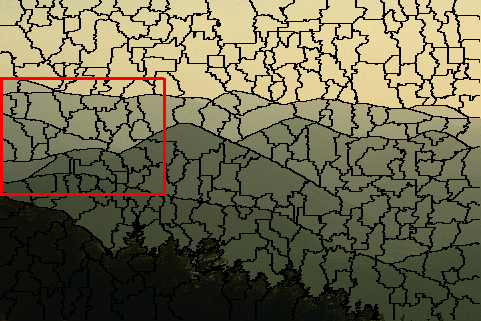}&
\includegraphics[height=\ppp,width=\wwwh]{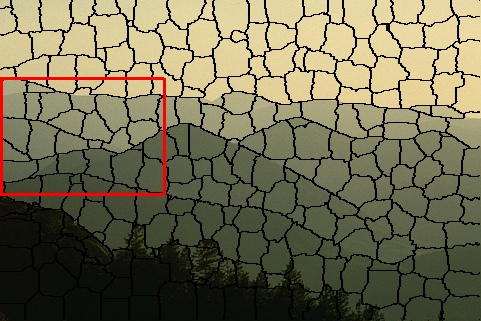}&
\includegraphics[height=\ppp,width=\wwwh]{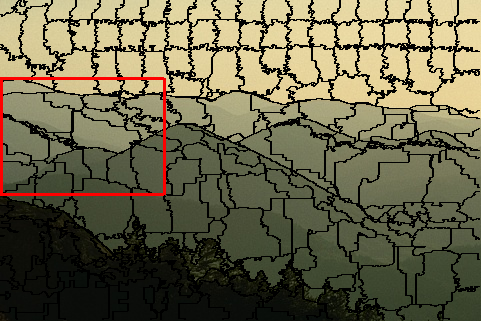}&
\includegraphics[height=\ppp,width=\wwwh]{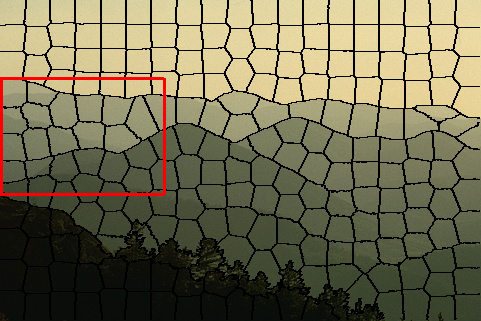}&
\includegraphics[height=\ppp,width=\wwwh]{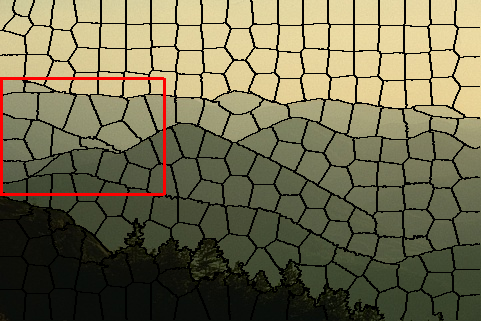}\\
\includegraphics[height=\ppp,width=\wwwh]{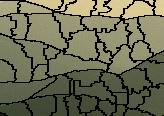}&
\includegraphics[height=\ppp,width=\wwwh]{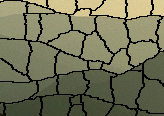}&
\includegraphics[height=\ppp,width=\wwwh]{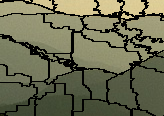}&
\includegraphics[height=\ppp,width=\wwwh]{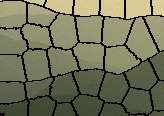}&
\includegraphics[height=\ppp,width=\wwwh]{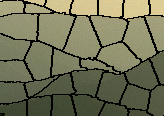}\\
\includegraphics[height=\ppp,width=\wwwh]{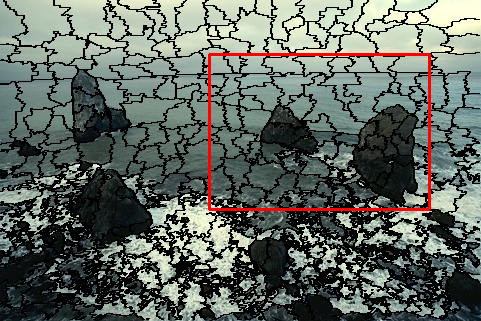}&
\includegraphics[height=\ppp,width=\wwwh]{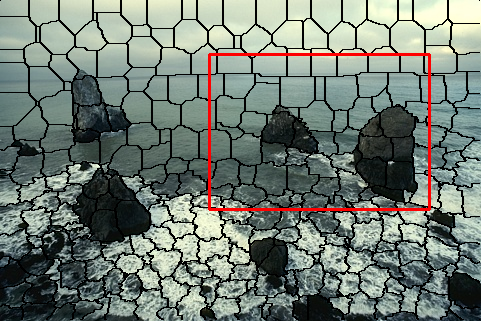}&
\includegraphics[height=\ppp,width=\wwwh]{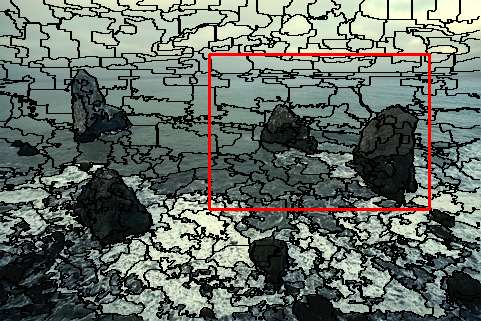}&
\includegraphics[height=\ppp,width=\wwwh]{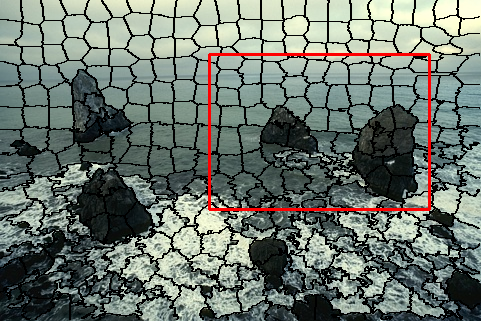}&
\includegraphics[height=\ppp,width=\wwwh]{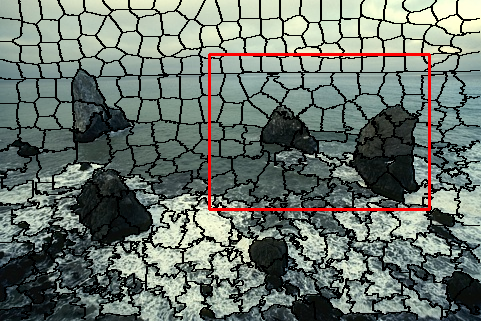}\\
\includegraphics[height=\ppp,width=\wwwh]{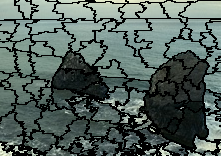}&
\includegraphics[height=\ppp,width=\wwwh]{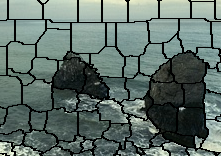}&
\includegraphics[height=\ppp,width=\wwwh]{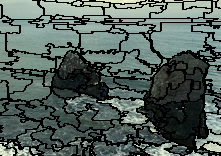}&
\includegraphics[height=\ppp,width=\wwwh]{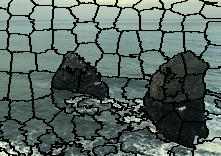}&
\includegraphics[height=\ppp,width=\wwwh]{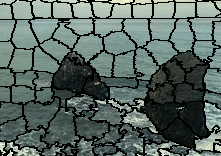}\\
\end{tabular}
} %\vspace{-0.1cm}
\caption{Comparison of decomposition results between SCALP and state-of-the-art superpixel methods on example images of the BSD %or $k=200$ superpixels.
} %\vspace{-0.1cm}
\label{fig:soa_images}
\end{figure*}

\section{Conclusion}

 In this work, we propose a generalization  of the superpixel clustering framework of \cite{achanta2012},
  by considering image feature and contour intensity on the linear path from the pixel to the superpixel barycenter.
  The contour prior information
  enhances the  adherence to the object boundaries.
  
  The proposed SCALP method provides superpixels of more regular shape, 
  according to the compactness measure.
  SCALP also obtains state-of-the-art results, 
  outperforming \cite{achanta2012} on superpixel metrics, and
  obtains the higher 
  $F$-measure
  among the compared methods.
  Finally, our fast integration of the contour prior within the framework enables to obtain the decomposition in a limited computational time.

Future works will focus on SCALP adaptation to supervoxel decomposition,
for video and 3D image processing.

\section*{Acknowledgment}
This study has been carried out with financial support from the French 
State, managed by the French National Research Agency (ANR) in the  
frame of the Investments for the future Program IdEx Bordeaux 
(ANR-10-IDEX-03-02), Cluster of excellence CPU and TRAIL (HR-DTI
ANR-10-LABX-57).

\bibliographystyle{IEEEtran}
\bibliography{ICPR}

\begin{thebibliography}{10}
\providecommand{\url}[1]{#1}
\csname url@samestyle\endcsname
\providecommand{\newblock}{\relax}
\providecommand{\bibinfo}[2]{#2}
\providecommand{\BIBentrySTDinterwordspacing}{\spaceskip=0pt\relax}
\providecommand{\BIBentryALTinterwordstretchfactor}{4}
\providecommand{\BIBentryALTinterwordspacing}{\spaceskip=\fontdimen2\font plus
\BIBentryALTinterwordstretchfactor\fontdimen3\font minus
  \fontdimen4\font\relax}
\providecommand{\BIBforeignlanguage}[2]{{%
\expandafter\ifx\csname l@#1\endcsname\relax
\typeout{** WARNING: IEEEtran.bst: No hyphenation pattern has been}%
\typeout{** loaded for the language `#1'. Using the pattern for}%
\typeout{** the default language instead.}%
\else
\language=\csname l@#1\endcsname
\fi
#2}}
\providecommand{\BIBdecl}{\relax}
\BIBdecl

\bibitem{arbelaez2011}
P.~Arbelaez, M.~Maire, C.~Fowlkes, and J.~Malik, ``Contour detection and
  hierarchical image segmentation,'' \emph{PAMI}, vol.~33, no.~5, pp. 898--916,
  2011.

\bibitem{kae2013}
A.~Kae, K.~Sohn, H.~Lee, and E.~Learned-Miller, ``Augmenting {CRF}s with
  {B}oltzmann machine shape priors for image labeling,'' in \emph{CVPR}, 2013,
  pp. 2019--2026.

\bibitem{fulkerson2009}
B.~Fulkerson, A.~Vedaldi, and S.~Soatto, ``Class segmentation and object
  localization with superpixel neighborhoods,'' in \emph{ICCV}, 2009, pp.
  670--677.

\bibitem{gould2008}
S.~Gould, J.~Rodgers, D.~Cohen, G.~Elidan, and D.~Koller, ``Multi-class
  segmentation with relative location prior,'' \emph{IJCV}, vol.~80, no.~3, pp.
  300--316, 2008.

\bibitem{gould2014}
S.~Gould, J.~Zhao, X.~He, and Y.~Zhang, ``Superpixel graph label transfer with
  learned distance metric,'' in \emph{ECCV}, 2014, pp. 632--647.

\bibitem{vandenbergh2012}
M.~Van~den Bergh, X.~Boix, G.~Roig, B.~de~Capitani, and L.~Van~Gool, ``{SEEDS}:
  Superpixels extracted via energy-driven sampling,'' in \emph{ECCV}, 2012, pp.
  13--26.

\bibitem{wang2011}
S.~Wang, H.~Lu, F.~Yang, and M.~H. Yang, ``Superpixel tracking,'' in
  \emph{ICCV}, 2011, pp. 1323--1330.

\bibitem{martin2001}
D.~Martin, C.~Fowlkes, D.~Tal, and J.~Malik, ``A database of human segmented
  natural images and its application to evaluating segmentation algorithms and
  measuring ecological statistics,'' in \emph{ICCV}, vol.~2, 2001, pp.
  416--423.

\bibitem{vincent91}
L.~Vincent and P.~Soille, ``Watersheds in digital spaces: an efficient
  algorithm based on immersion simulations,'' \emph{PAMI}, vol.~13, no.~6, pp.
  583--598, 1991.

\bibitem{comaniciu2002}
D.~Comaniciu and P.~Meer, ``Mean shift: a robust approach toward feature space
  analysis,'' \emph{PAMI}, vol.~24, no.~5, pp. 603--619, 2002.

\bibitem{vedaldi2008}
A.~Vedaldi and S.~Soatto, ``Quick shift and kernel methods for mode seeking,''
  in \emph{ECCV}, 2008, pp. 705--718.

\bibitem{felzenszwalb2004}
P.~F. Felzenszwalb and D.~P. Huttenlocher, ``Efficient graph-based image
  segmentation,'' \emph{IJCV}, vol.~59, no.~2, pp. 167--181, 2004.

\bibitem{levinshtein2009}
A.~Levinshtein, A.~Stere, K.~N. Kutulakos, D.~J. Fleet, S.~J. Dickinson, and
  K.~Siddiqi, ``Turbopixels: Fast superpixels using geometric flows,''
  \emph{PAMI}, vol.~31, no.~12, pp. 2290--2297, 2009.

\bibitem{veksler2010}
O.~Veksler, Y.~Boykov, and P.~Mehrani, ``Superpixels and supervoxels in an
  energy optimization framework,'' in \emph{ECCV}, 2010, pp. 211--224.

\bibitem{liu2011}
M.~Y. Liu, O.~Tuzel, S.~Ramalingam, and R.~Chellappa, ``Entropy rate superpixel
  segmentation,'' in \emph{CVPR}, 2011, pp. 2097--2104.

\bibitem{machairas2015}
V.~Machairas, M.~Faessel, D.~Cárdenas-Peña, T.~Chabardes, T.~Walter, and
  E.~Decencière, ``Waterpixels,'' \emph{TIP}, vol.~24, no.~11, pp. 3707--3716,
  2015.

\bibitem{achanta2012}
R.~Achanta, A.~Shaji, K.~Smith, A.~Lucchi, P.~Fua, and S.~Süsstrunk, ``{SLIC}
  superpixels compared to state-of-the-art superpixel methods,'' \emph{PAMI},
  vol.~34, no.~11, pp. 2274--2282, 2012.

\bibitem{li2015}
Z.~Li and J.~Chen, ``Superpixel segmentation using linear spectral
  clustering,'' in \emph{CVPR}, 2015, pp. 1356--1363.

\bibitem{zhang2016}
Y.~Zhang, X.~Li, X.~Gao, and C.~Zhang, ``A simple algorithm of superpixel
  segmentation with boundary constraint,'' \emph{TCSVT}, no.~99, 2016.

\bibitem{arbelaez2009}
P.~Arbelaez, M.~Maire, C.~Fowlkes, and J.~Malik, ``From contours to regions: An
  empirical evaluation,'' in \emph{CVPR}, 2009, pp. 2294--2301.

\bibitem{mori2004}
G.~Mori, X.~Ren, A.~A. Efros, and J.~Malik, ``Recovering human body
  configurations: combining segmentation and recognition,'' in \emph{CVPR},
  vol.~2, 2004, pp. 326--333.

\bibitem{shi2000}
J.~Shi and J.~Malik, ``Normalized cuts and image segmentation,'' \emph{PAMI},
  vol.~22, no.~8, pp. 888--905, 2000.

\bibitem{moore2008}
A.~P. Moore, S.~J.~D. Prince, J.~Warrell, U.~Mohammed, and G.~Jones,
  ``Superpixel lattices,'' in \emph{CVPR}, 2008, pp. 1--8.

\bibitem{martin2004}
D.~R. Martin, C.~C. Fowlkes, and J.~Malik, ``Learning to detect natural image
  boundaries using local brightness, color, and texture cues,'' \emph{PAMI},
  vol.~26, no.~5, pp. 530--549, 2004.

\bibitem{bresenham1965}
J.~E. Bresenham, ``Algorithm for computer control of a digital plotter,''
  \emph{IBM Syst. J.}, vol.~4, no.~1, pp. 25--30, 1965.

\bibitem{zeng2011}
G.~Zeng, P.~Wang, J.~Wang, R.~Gan, and H.~Zha, ``Structure-sensitive
  superpixels via geodesic distance,'' in \emph{ICCV}, 2011, pp. 447--454.

\bibitem{maire2008}
M.~Maire, P.~Arbelaez, C.~Fowlkes, and J.~Malik, ``Using contours to detect and
  localize junctions in natural images,'' in \emph{CVPR}, 2008, pp. 1--8.

\bibitem{xiaofeng2012}
R.~Xiaofeng and L.~Bo, ``Discriminatively trained sparse code gradients for
  contour detection,'' in \emph{NIPS}, 2012, pp. 584--592.

\bibitem{dollar2013}
P.~Doll\'{a}r and C.~L. Zitnick, ``Structured forests for fast edge
  detection,'' in \emph{ICCV}, 2013, pp. 1841--1848.

\bibitem{schick2012}
A.~Schick, M.~Fischer, and R.~Stiefelhagen, ``Measuring and evaluating the
  compactness of superpixels,'' in \emph{ICPR}, 2012, pp. 930--934.

\end{thebibliography}

\end{document}